\documentclass{article}

\usepackage{microtype}
\usepackage{graphicx}
\usepackage{subcaption}
\usepackage{booktabs}

\usepackage{hyperref}
\usepackage{xurl}

\ifdefined\isanon
  \usepackage{icml2026}
\else
  \usepackage[preprint]{icml2026}
\fi

\usepackage{amsmath}
\usepackage{amssymb}
\usepackage{mathtools}
\usepackage{amsthm}

\usepackage[capitalize,noabbrev]{cleveref}
\usepackage{placeins}

\theoremstyle{plain}

\theoremstyle{definition}

\theoremstyle{remark}

\usepackage[textsize=tiny]{todonotes}

\icmltitlerunning{Evidence for feature-specific error correction in LLMs}

\begin{document}

\twocolumn[
  \icmltitle{Evidence for feature-specific error correction in LLMs}

  \icmlsetsymbol{equal}{*}

  \begin{icmlauthorlist}
    \icmlauthor{Francisco Ferreira da Silva}{pivotal}
    \icmlauthor{Stefan Heimersheim}{}
  \end{icmlauthorlist}

  \icmlaffiliation{pivotal}{Pivotal Research}

  \icmlcorrespondingauthor{Francisco Ferreira da Silva}{franhfsilva@gmail.com}

  \icmlkeywords{Feature Geometry, Error Correction, Computation in Superposition}

  \vskip 0.3in
]

\printAffiliationsAndNotice{}

\begin{abstract}
Understanding the features of large language models (LLMs) is a central goal of interpretability.
LLMs are commonly assumed to use superposition to represent more features than they have dimensions.
They may not only represent features in superposition but also perform computation in superposition.
Theory predicts that computing in superposition requires error correction that privileges feature directions over generic ones, but this prediction has not been tested empirically.
We propose an empirical test of error correction in LLMs based on activation perturbations.
Perturbing residual-stream activations, we find that they are robust to small perturbations---forming activation plateaus consistent with error correction---but less robust along candidate feature directions ("pure" directions, constructed from contrastive prompt pairs) than along mixtures of two such directions, indicating that the pure directions are privileged.
We quantify this privilegedness by modeling the perturbation effect as a function of the $L^p$-norm of its decomposition into feature components.
For $p=2$ the response is a quadratic form with at most as many nonzero eigenvalues as the residual-stream dimension, which cannot privilege the many feature directions superposition requires.
$p>2$ lifts this constraint and is consistent with feature-specific error correction.
We find $p>2$ for contrastive, MELBO, and SAE-decoder directions, and $p\approx2$ for random and PCA directions (controls).
These results replicate across Gemma-2-9B, Qwen3-1.7B, Llama-3.1-8B, Mistral-7B-v0.3, Aya-Expanse-8B, and Yi-1.5-9B.
We further validate our method on a toy model of error correction with known ground-truth features, recovering $p>2$ for true feature directions, degrading toward $2$ as we rotate away from them.\ifdefined\isanon{\renewcommand{\thefootnote}{\ensuremath{\dagger}}\footnote{Code: \url{https://anonymous.4open.science/r/fsec-anonymous-5C9D/}}\addtocounter{footnote}{-1}}\else{\renewcommand{\thefootnote}{\ensuremath{\dagger}}\footnote{Code: \url{https://github.com/FranciscoHS/fsec-paper}}\addtocounter{footnote}{-1}}\fi
\end{abstract}

\section{Introduction}
\label{sec:introduction}
Representations in large language models (LLMs) are poorly understood.
It is commonly assumed that LLMs make use of superposition~\cite{elhage2022toy} to represent more concepts than they have dimensions available, and potentially to compute in superposition (CiS)~\cite{hanni2024mathematical,adler2024complexity,olah2025interference}.
However, we have only indirect evidence for superposition, chiefly the success of sparse autoencoders~\cite{cunningham2023sparse, gao2024scaling, templeton2026scaling} at extracting interpretable directions, and only theoretical 
evidence for CiS~\cite{hanni2024mathematical}.

An empirical prediction of CiS is that neural networks must correct interference noise while preserving feature signal~\cite{hanni2024mathematical}.
This requires networks to be less sensitive to perturbations along non-feature directions than along feature directions.
We name this property feature-specific error correction (FSEC).
While we cannot rule FSEC out or in without ground-truth feature directions, we can still ask whether FSEC-like behaviour occurs for any directions: if the model's error correction privileges certain directions, those directions are candidate features, and we can detect them by measuring sensitivity, requiring only a generic input to perturb rather than feature-specific labeled data.

We provide the first empirical evidence of FSEC, showing that the robustness of LLM activations to perturbations privileges certain candidate feature directions over others.
Concretely, we perturb residual-stream activations at early layers and measure the downstream response as a function of perturbation direction and magnitude.
We construct candidate feature directions via contrastive means for a variety of concepts, including languages, programming languages, gender, sentiment, registers, and verb tenses.
Across Gemma-2-9B~\cite{team2024gemma}, Qwen3-1.7B~\cite{yang2025qwen3}, Llama-3.1-8B~\cite{grattafiori2024llama}, Mistral-7B-v0.3~\cite{Jiang2023Mistral7}, Aya-Expanse-8B~\cite{dang2024aya}, and Yi-1.5-9B~\cite{young2024yi}, we find that contrastive feature directions elicit a stronger downstream response than mixtures thereof, consistent with FSEC that privileges feature directions while suppressing interference along non-feature directions.
We formalize this by modeling the LLMs' response to perturbation as an $L^p$ norm of the perturbation's decomposition into candidate feature directions.
The $p=2$ case reduces to a basis-invariant quadratic form, meaning no choice of basis is privileged; $p>2$ breaks this invariance, indicating that the candidate pure feature directions are more sensitive than their mixtures, as predicted by FSEC.
We measure $p \approx 2.3$ for contrastive directions across models.
MELBO~\cite{mack2024melbo} and SAE directions, which also aim to recover model features, likewise yield $p > 2$, albeit with smaller values.
PCA and random directions do not, consistent with the interpretation that $p > 2$ reflects alignment with the model's features.
We validate this methodology in a toy model of error correction with ground-truth features (Section~\ref{sec:toy_model}), confirming that $p$ degrades toward $2$ as directions are misaligned with the true features.

Our contributions are:
\begin{enumerate}
\item We propose feature-specific error correction (FSEC) as a test of computation in superposition: we model the LLM's response to a perturbation as a function of the $L^p$-norm of the perturbation's decomposition into candidate feature directions, and FSEC predicts $p > 2$.
\item We find evidence of FSEC for three types of candidate feature directions---contrastive, SAE-decoder, and MELBO---each with $p > 2$, and show the contrastive result replicates across six LLMs from different families.
PCA and random directions yield $p \approx 2$, consistent with them not being privileged.
\item We also show that FSEC occurs in a toy model of error correction~\cite{vaintrob2026tale}.
\end{enumerate}

\section{Related Work}
\textbf{Activation plateaus.} Prior work has established that in-distribution activations of LLMs are resistant to perturbations~\cite{heimersheim2024plateaus,janiak2024characterizing,shinkle2025plateaus}, a phenomenon known as \textit{activation plateaus}.
We introduce a novel measurement of the activation plateau boundary geometry, and identify its connection to FSEC.

\textbf{Direction-dependent sensitivity for feature finding.} Prior work has exploited the fact that LLMs have direction-dependent sensitivity to perform unsupervised optimizations for directions maximizing downstream response, resulting in interpretable steering vectors, including MELBO~\cite{mack2024melbo,mack2024deepcausaltranscoding}.
We make use of the same phenomenon to empirically probe error correction, and apply our analysis to directions obtained in this way (Section~\ref{sec:ablations}).
Unlike this line of work, our analysis adds the novel study of the sensitivity geometry and its connection to error correction.

\textbf{Error correction for computation in superposition.} \citet{hanni2024mathematical} argue theoretically that computation in superposition requires error correction that privileges feature directions.
We provide empirical evidence in favor of superposition and error correction occurring in LLMs.

\section{Methodology}
\label{sec:methods}
We probe error correction in LLMs by perturbing residual stream activations and measuring the downstream response. In-distribution activations are robust to small perturbations, a phenomenon known as \textit{activation plateaus}~\cite{heimersheim2024plateaus,shinkle2025plateaus}. This robustness is direction-dependent, i.e., the model is more sensitive to perturbations along some directions than others. Two lines of evidence suggest feature directions in particular are privileged: empirically, prior work recovers interpretable directions by optimizing for sensitivity, the inverse of robustness~\cite{mack2024melbo,mack2024deepcausaltranscoding}; and theoretically, FSEC predicts that feature directions are privileged. We test this by comparing the downstream response along candidate feature directions to the response along non-feature baselines.

In all experiments, we perturb the residual stream at an early layer $\ell$ (default $\ell=2$) and measure the downstream response at the second-to-last layer, maximizing the number of intervening layers.
This is because activation plateaus are known to be more pronounced the greater the distance between perturbation and measurement~\cite{shinkle2025plateaus}.
We avoid the last residual stream layer, which is known to behave atypically.
The downstream response is computed by patching the perturbed activation $\textbf{a}(\alpha)$ back into the model~\cite{meng2022locating,heimersheim2024use}, performing a forward pass, and taking the $L^2$ distance between the perturbed and unperturbed residual streams at the measurement layer.
We show in Section~\ref{sec:ablations} that our results are robust to varying both the perturbation and measurement layers, and hold also when measuring cosine distance or KL-divergence in the logits.

To distinguish directional effects from those of magnitude, we follow prior perturbation analyses of activation plateaus~\cite{heimersheim2024plateaus} and perturb by rotating the activation vector $\textbf{a}$ towards the perturbation direction $\textbf{d}$ while keeping the activation's norm constant.
We refer to this as a \textit{norm-matched perturbation}.
Concretely, a perturbation of angle $\alpha$ of $\textbf{a}$ toward $\textbf{d}$ is:
\begin{equation}
\label{eq:norm-matched}
\textbf{a}(\alpha)=\cos(\alpha)\textbf{a}+\sin(\alpha)\left\lVert \textbf{a} \right\rVert \frac{\textbf{d}_{\perp}}{\left\lVert \textbf{d}_{\perp} \right\rVert},
\end{equation}
where $\textbf{d}_{\perp}$ is the component of $\textbf{d}$ that is orthogonal to $\textbf{a}$.
We typically perturb at the last token position, but show in Section~\ref{sec:ablations} that our results are robust to this choice.

We quantify the model's sensitivity along a given direction as the \textit{plateau-breaking angle}, the smallest perturbation angle for which the downstream response exceeds a threshold $T$.
We set $T$ per direction-pair, at a level that is guaranteed to be crossed by both single-axis sweeps.\footnote{We initially tried a single global threshold derived from random directions---a fraction $f$ of the median, over a set of isotropic random unit directions, of the single-axis plateau height each one reaches---but abandoned it. Direction sensitivities span a wide range (up to a factor of ${\sim}70$ between the least and most responsive directions for the KL-divergence response metric), so a global threshold is either never reached by the least responsive directions or crossed only at very small angles by the most responsive ones, leaving most superellipse fits ill-defined. Setting $T$ per pair from each pair's own single-axis maxima guarantees both axes cross and resolves this. We record this here for ease of reproducibility.}
We aggregate the downstream response across a fixed set of $N = 30$ inputs (\textit{anchors}) $\{\mathbf{x}_n\}_{n=1}^{N}$---last-token residual-stream activations of 5-token FineWeb prompts~\cite{penedo2024fineweb}---by taking the median over anchors.
Writing $L^2(\mathbf{x}_n, \alpha; \mathbf{d})$ for the response of anchor $\mathbf{x}_n$ when perturbed by angle $\alpha$ along direction $\mathbf{d}$, the single-axis response curve is
\begin{equation}
  \label{eq:median-response}
  L^2(\alpha; \mathbf{d}) = \operatorname*{median}_{n=1,\dots,N} L^2(\mathbf{x}_n, \alpha; \mathbf{d}).
\end{equation}
$\max_\alpha L^2(\alpha; \mathbf{d}_i)$ is then the largest median response attained when sweeping along $\mathbf{d}_i$ alone.
For a pair $(\mathbf{d}_1, \mathbf{d}_2)$ we set the threshold
\begin{equation}
  T = f \cdot \min\!\big(\max_\alpha L^2(\alpha; \mathbf{d}_1),\; \max_\alpha L^2(\alpha; \mathbf{d}_2)\big),
\end{equation}
a fraction $f$ of the smaller of the two single-axis maxima.
Taking the smaller maximum ensures both axes reach $T$, so the single-axis plateau-breaking angles $\alpha_1, \alpha_2$ that calibrate the fit are always defined.
Throughout we use $f = 0.5$.
We show in Section~\ref{sec:ablations} that our results are robust to varying $T$ within half-to-double its nominal value.

We perform perturbations along six types of directions: contrastive~\cite{panickssery2023steering,turner2023steering}, MELBO~\cite{mack2024melbo}, SAE latents~\cite{lieberum2024gemma}, PCA directions, random, and random-difference directions.
The first three are all candidate feature directions, while PCA, random, and random-difference directions function as non-feature baselines.
PCA directions are computed by performing PCA on the residual-stream activations arising from a randomly selected sample of 10000 5-token-long FineWeb inputs~\cite{penedo2024fineweb}.
Random directions are sampled isotropically on the unit sphere,
$\hat{\mathbf{d}}_{\text{rand}} \sim \mathrm{Unif}(\mathcal{S}^{d-1})$.

Contrastive directions are constructed as the difference of mean
activations between two sets of $P\geq30$ matched prompt pairs
$\mathcal{P}^+, \mathcal{P}^-$ differing along a single concept (e.g., for gender, $\mathcal{P}^+ \ni$ ``He ran home'' is paired with $\mathcal{P}^- \ni$ ``She ran home'')~\cite{panickssery2023steering,turner2023steering}.
The prompts are LLM generated and human verified, and are available in our code release.
Letting $\mathbf{a}(s) \in \mathbb{R}^d$
denote the activation at the perturbation layer $\ell$ for prompt $s$,
\begin{align}
  \mathbf{d}_{\text{contrast}} \;&=\;
  \frac{1}{P}\sum_{i=1}^{P} \big(\mathbf{a}(s_i^+) - \mathbf{a}(s_i^-)\big),
  \\
  \hat{\mathbf{d}}_{\text{contrast}} \;&=\; \mathbf{d}_{\text{contrast}} / \|\mathbf{d}_{\text{contrast}}\|_2.
\end{align}

Random-difference directions apply this same construction to randomly paired activations rather than concept-matched ones, giving a control that is matched to contrastive directions in everything except semantic content.
We average $P = 30$ differences between pairs $(s_i, s_i')$ drawn at random from the same 10000-input FineWeb sample used for PCA, and normalize:
\begin{align}
\mathbf{d}_{\text{rand-diff}}\;&=\; \frac{1}{P}\sum_{i=1}^{P} \big(\mathbf{a}(s_i) - \mathbf{a}(s_i')\big),
\\
\hat{\mathbf{d}}_{\text{rand-diff}} \;&=\; \mathbf{d}_{\text{rand-diff}} / \|\mathbf{d}_{\text{rand-diff}}\|_2.
\end{align}
Unlike contrastive directions, the paired prompts share no concept, so we do not expect them to be features. Unlike isotropic random directions, $\hat{\mathbf{d}}_{\text{rand-diff}}$ consist of differences of activations, and hence inherit their covariance.
This makes it our most stringent non-feature baseline: it differs from a contrastive direction only in the absence of concept-coherent pairing.
We construct 40 such directions.

MELBO directions~\cite{mack2024melbo} are constructed by optimising a unit-norm direction $\hat{\mathbf{d}}$ at the perturbation layer to maximise the $L^2$ distance between perturbed and unperturbed activations at a downstream layer.
Unlike contrastive directions, this requires no labelled prompt pairs as the directions are found unsupervised, and have been shown to produce interpretable steering when applied at sufficient magnitude.

SAE latents are decoder columns of a sparse autoencoder trained on the model's residual stream activations~\cite{bricken2023monosemanticity, cunningham2023sparse}.
Each column is a candidate feature direction under the SAE's decomposition.
We select the top-33 most active SAE latents in the same sample of 10000 FineWeb inputs used for PCA directions.
We do this for Gemma-2-9B only and use Gemma Scope~\cite{lieberum2024gemma} (the width-16k residual-stream SAE at layer~2).

We also perform perturbations toward combinations of two directions $\mathbf{d}_1, \mathbf{d}_2$.
We first restrict them to the tangent space at $\hat{\mathbf{a}} = \mathbf{a}/\lVert\mathbf{a}\rVert$ via the projector $I - \hat{\mathbf{a}}\hat{\mathbf{a}}^\top$, and orthonormalize (Gram–Schmidt) to obtain unit vectors $\hat{\mathbf{d}}_1^{\perp}, \hat{\mathbf{d}}_2^{\perp}$ satisfying $\hat{\mathbf{a}} \cdot \hat{\mathbf{d}}_i^{\perp} = 0$ and $\hat{\mathbf{d}}_1^{\perp} \cdot \hat{\mathbf{d}}_2^{\perp} = 0$. We then form the unit tangent direction
\begin{equation}
  \hat{\mathbf{d}}(\varphi) \;=\; \cos\varphi\, \hat{\mathbf{d}}_1^{\perp} + \sin\varphi\, \hat{\mathbf{d}}_2^{\perp}, \qquad \varphi \in [0, \pi/2],
\end{equation}
which interpolates between $\hat{\mathbf{d}}_1^{\perp}$ at $\varphi=0$ and $\hat{\mathbf{d}}_2^{\perp}$ at $\varphi=\pi/2$.
Perturbing $\mathbf{a}$ by angle $\alpha$ toward $\hat{\mathbf{d}}(\varphi)$ then proceeds exactly as in the single-direction case (Equation~\ref{eq:norm-matched}),
\begin{equation}
  \label{eq:norm-matched-phi}
  \mathbf{a}(\alpha, \varphi) \;=\; \cos\alpha\, \mathbf{a} + \sin\alpha\, \lVert\mathbf{a}\rVert\, \hat{\mathbf{d}}(\varphi).
\end{equation}

We observe behavior (see Section~\ref{sec:results}) that appears consistent with the downstream response depending on the perturbation's projections onto privileged directions, raised to a common power $p$.
To formalize this, we model the downstream response $R$ to a perturbation vector $\mathbf{v}$ as
\begin{equation}
\label{eq:response-model}
R(\mathbf{v}) = F\!\left(\sum_i w_i\,\bigl|\hat{\mathbf{f}}_i^\top \mathbf{v}\bigr|^p\right),
\end{equation}
where $\{\hat{\mathbf{f}}_i\}$ are unit directions, the weights $w_i \geq 0$ are free parameters capturing each direction's sensitivity, and $F$ is a scalar function.

When we perturb along two directions simultaneously and vary their relative weighting via the mixing angle $\varphi$, the plateau-breaking angle becomes a function of $\varphi$; plotted in suitable coordinates, this function traces out a superellipse (see Figure~\ref{fig:iso-plateau-boundary}).
Taking $\hat{\mathbf{f}}_1 = \hat{\mathbf{d}}_1^{\perp}$ and $\hat{\mathbf{f}}_2 = \hat{\mathbf{d}}_2^{\perp}$, the inner sum collapses to two terms provided the remaining $\hat{\mathbf{f}}_j$ contribute negligibly along $\hat{\mathbf{d}}(\varphi)$.
This is expected under superposition, where features are packed nearly orthogonally~\cite{elhage2022toy}, so the projections of $\hat{\mathbf{d}}(\varphi)$ onto other directions $\hat{\mathbf{f}}_j$ are small; the quality of the superellipse fits in Section~\ref{sec:results} confirms the approximation.
For the norm-matched perturbation of Equation~\ref{eq:norm-matched-phi}, the displacement is $\mathbf{v} = \mathbf{a}(\alpha,\varphi) - \mathbf{a} = \sin\alpha\,\lVert\mathbf{a}\rVert\,\hat{\mathbf{d}}(\varphi) - (1-\cos\alpha)\,\mathbf{a}$. The term $-(1-\cos\alpha)\,\mathbf{a}$ lies along $\mathbf{a}$ and is therefore orthogonal to each $\hat{\mathbf{f}}_i = \hat{\mathbf{d}}_i^{\perp}$, so it drops out of every projection, $\hat{\mathbf{f}}_i^\top \mathbf{v} = \sin\alpha\,\lVert\mathbf{a}\rVert\,(\hat{\mathbf{f}}_i^\top \hat{\mathbf{d}}(\varphi))$; only the tangential component contributes, so
\begin{equation}
R = F\!\left(\sin^p\alpha\,\lVert\mathbf{a}\rVert^p \bigl[w_1 \cos^p\varphi + w_2 \sin^p\varphi\bigr]\right).
\end{equation}
The plateau breaks when $F$'s argument crosses a fixed level $\xi^*$ (the value of the argument at which the response reaches the threshold $T$), giving the implicit equation for the plateau-breaking angle $\alpha(\varphi)$:
\begin{equation}
\sin^p\alpha(\varphi)\,\lVert\mathbf{a}\rVert^p \bigl[w_1 \cos^p\varphi + w_2 \sin^p\varphi\bigr] = \xi^*.
\end{equation}
The single-axis sweeps (at $\varphi = 0$ and $\varphi = \pi/2$) calibrate the weights at this same threshold: $\varphi = 0$ gives $w_1\,\lVert\mathbf{a}\rVert^p = \xi^*/\sin^p\alpha_1$, and $\varphi = \pi/2$ gives $w_2\,\lVert\mathbf{a}\rVert^p = \xi^*/\sin^p\alpha_2$, where $\alpha_1, \alpha_2$ are the plateau-breaking angles measured along $\hat{\mathbf{d}}_1^{\perp}$ and $\hat{\mathbf{d}}_2^{\perp}$ individually.
We calibrate on the single-axis sweeps because each isolates one weight; this is a choice of convenience rather than a requirement.
Substituting and dividing through by $\xi^*$ eliminates the unknown threshold and the activation norm, yielding
\begin{equation}
\left(\frac{\sin\alpha(\varphi)\cos\varphi}{\sin\alpha_1}\right)^{\!p} + \left(\frac{\sin\alpha(\varphi)\sin\varphi}{\sin\alpha_2}\right)^{\!p} = 1.
\end{equation}
This means that the plateau-breaking angles define a superellipse of exponent $p$ in the normalised coordinates
$(\sin\alpha(\varphi)\,\cos\varphi/\sin\alpha_1,\;
  \sin\alpha(\varphi)\,\sin\varphi/\sin\alpha_2)$.
We estimate $p$ by minimizing the squared residual of this equation
over the measured
$(\sin\alpha(\varphi)\,\cos\varphi,\;
  \sin\alpha(\varphi)\,\sin\varphi)$ points.

\section{Results}
\label{sec:results}
We begin by illustrating our measurement procedure on a concrete example.
\begin{figure}[!t]
  \centering
  \includegraphics[width=\columnwidth]{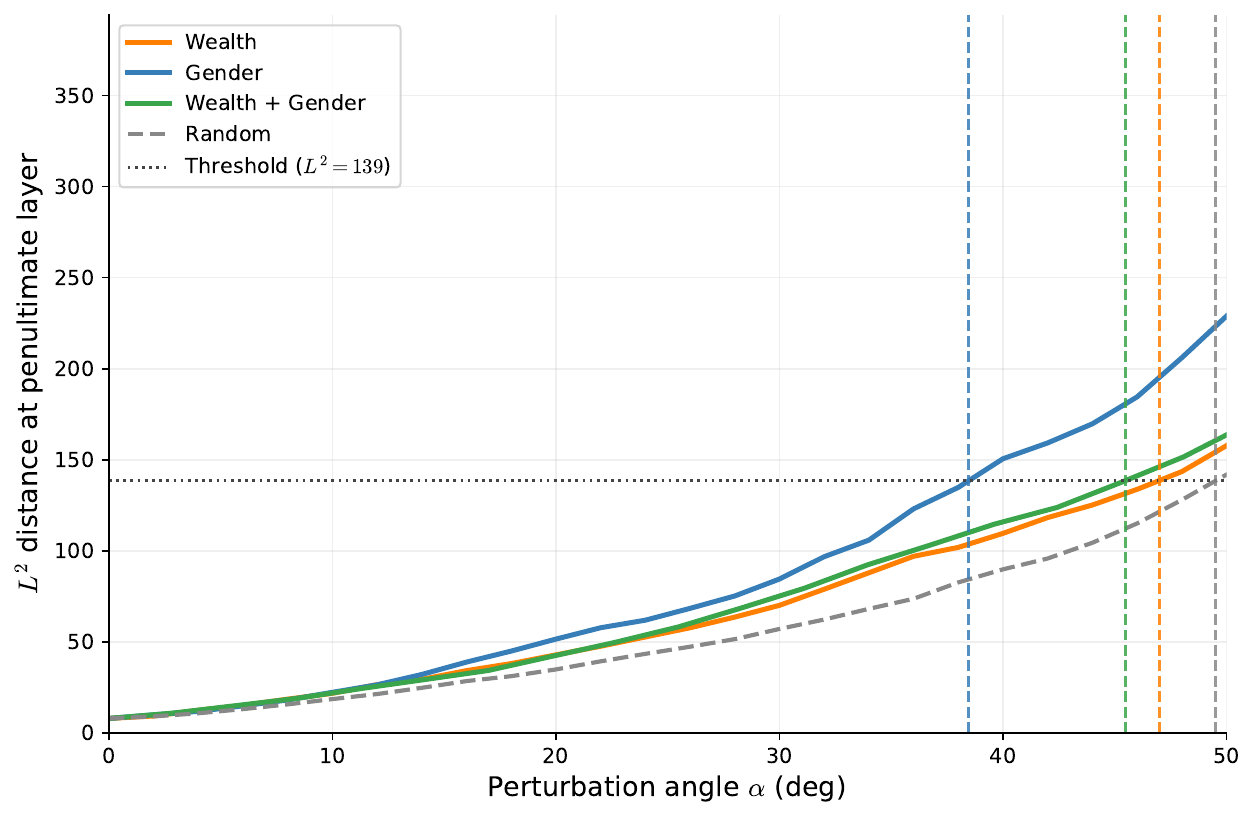}
  \caption{\textbf{Measuring plateau-breaking angles.}
    Downstream response as a function of perturbation angle for two contrastive directions (Wealth, Gender) and an equal combination of both, at Gemma-2-9B layer 2, illustrating how plateau-breaking angles are extracted.
    The plateau-breaking angle is the angle at which the downstream $L^2$ distance first exceeds the threshold (here $T\approx139$).
    The grey dashed ``Random'' curve shows the median response across perturbations along 10 isotropic random unit directions.
    All curves are medians over 30 FineWeb anchor prompts; we omit uncertainty bands because the absolute $L^2$ scale varies substantially from anchor to anchor in a way that is shared across all curves and largely cancels in within-anchor comparisons.
    The within-anchor gap between each feature direction and the random baseline is nonetheless robust: at every angle, the feature curve exceeds the random baseline for most anchors ($\geq\!63\%$), and the median feature-minus-random difference is significantly positive (its 95\% bootstrap confidence interval over anchors excludes zero).}
  \label{fig:plateau-breaking-measurement}
\end{figure}

\begin{figure}[!t]
  \centering
  \includegraphics[width=\columnwidth]{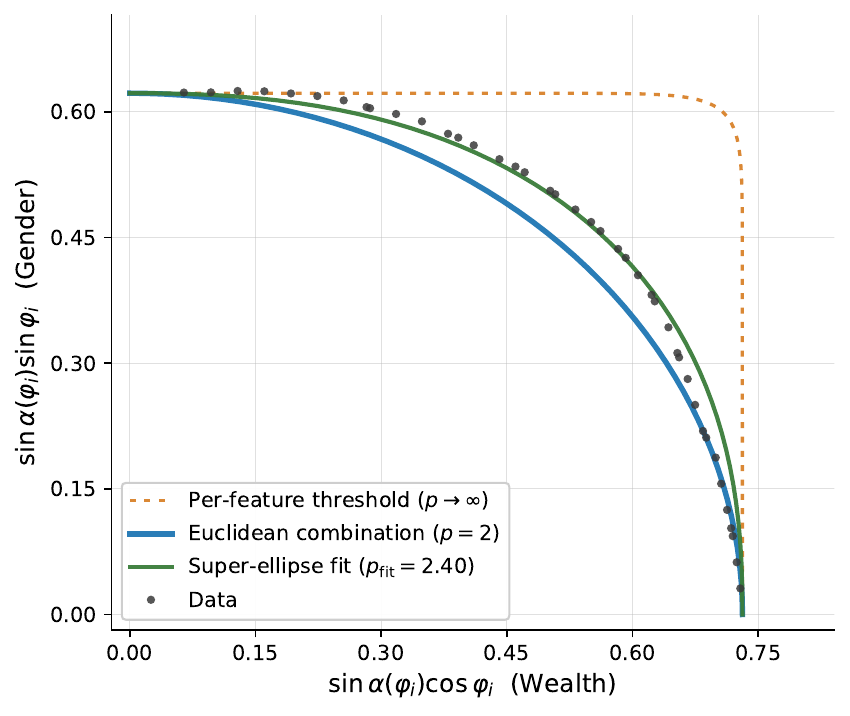}
  \caption{\textbf{Iso-plateau boundary.}
    Plateau-breaking angles for the Wealth $\times$ Gender pair at Gemma-2-9B layer 2 (per-pair threshold $T \approx 139$).
    The superellipse exponent is fit in the normalised coordinates $(\sin\alpha(\varphi)\cos\varphi/\sin\alpha_1,\; \sin\alpha(\varphi)\sin\varphi/\sin\alpha_2)$ of Section~\ref{sec:methods}.
    The boundary is well fit by a superellipse of exponent $p_{\mathrm{fit}}=2.40$ (fit residual $1.2\%$); $p > 2$ indicates these directions are privileged.}
  \label{fig:iso-plateau-boundary}
\end{figure}
Figure~\ref{fig:plateau-breaking-measurement} shows the downstream $L^2$ response in Gemma-2-9B when perturbing along the Wealth contrastive direction, the Gender contrastive direction, and an equal combination of the two.
The vertical dashed lines mark the plateau-breaking angle for each of the directions.
Repeating this measurement across a range of mixing angles $\varphi$ yields a plateau-breaking angle for each, which we plot in Figure~\ref{fig:iso-plateau-boundary}.
The resulting boundary is well fit by a superellipse with exponent $p_{\mathrm{fit}} = 2.40$.

\begin{figure*}
  \centering
  \includegraphics[width=0.92\textwidth]{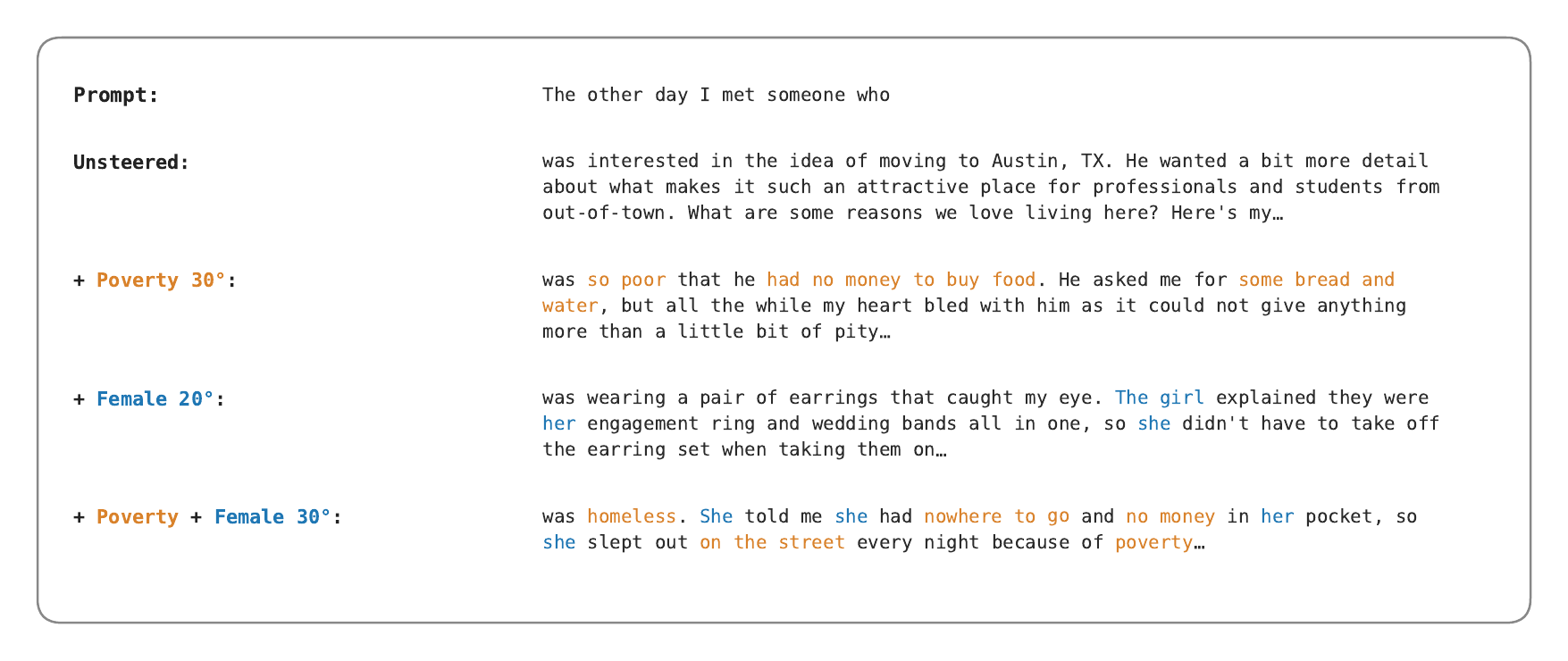}
  \caption{\textbf{Compositional steering at Gemma-2-9B layer 2.} Sample
    completions for the prompt ``The other day I met someone who'' under no
    steering, $+$Poverty alone, $+$Female alone, and the Poverty $+$ Female
    composite. Steering uses the contrastive Wealth and Gender directions
    (Appendix~\ref{app:directions}); each row shows a single pole---the
    low-wealth (poverty) pole of Wealth and the feminine pole of Gender.
    Highlighted spans: \textcolor[HTML]{D9822B}{orange} for poverty and
    \textcolor[HTML]{1F77B4}{blue} for feminine words.}
  \label{fig:composition}
\end{figure*}

To verify that combining contrastive directions in this way is meaningful, we show in Figure~\ref{fig:composition} that steering along these combinations produces interpretable compositional behaviour changes.
Concretely, steering along the Wealth direction (toward its low-wealth pole) produces poverty-themed completions from a neutral input, steering along the Gender direction produces feminine-coded completions, and steering along a combination of the two produces completions that are both poverty-themed and feminine-coded.
This supports the reasonableness of the combination-perturbation setup, though we note that interpretable steering alone does not establish that a direction corresponds to a model feature.

\subsection{Superellipse exponents across direction types}
\label{sec:exponents}
We now systematically characterise these patterns across directions, direction types, and models.
The superellipse exponent $p$ has a simple geometric meaning.
Perturbing along a single direction breaks the plateau at some angle; perturbing along a mixture of two directions may break it sooner, later, or at the same point, and which of these occurs is exactly what $p$ records.
If the model responds only to the overall size of a perturbation (its $L^2$ norm), then splitting it evenly between two directions is no different from concentrating it on one, and the plateau-breaking angles trace an ellipse ($p = 2$).
If instead the model is specially sensitive to the individual directions---so that a perturbation matters only insofar as it aligns with one of them---then an even mixture, which aligns only partially with each, must be pushed further before the plateau breaks.
The boundary then bulges outward, reaching a square ($p \to \infty$) in the limit where each direction has a fully independent threshold and the response fires only as the projection onto either one crosses it.
The exponent $p > 2$ measures how far toward this ``independent thresholds'' regime the model sits.
Each axis is normalised by its own plateau-breaking angle so that only the privileging of the pure directions over their mixtures---not their individual sensitivities---affects $p$.
This selectivity is exactly what error correction requires: staying responsive to individual features while suppressing generic mixtures of them.
We give the formal statement---that $p = 2$ cannot privilege feature directions under superposition whereas $p > 2$ can---in Section~\ref{sec:why-p-gt-2}.

We repeat the analysis across six direction types.
For contrastive directions, we form 318 pairs from 33 directions (14 binary semantic concepts, 10 natural languages, 9 programming languages), keeping only pairs with intra-pair cosine similarity below 0.1.
The filter is needed because our two-direction analysis orthogonalises the pair (Section~\ref{sec:methods}): if the two directions already have significant overlap, orthogonalising one against the other distorts it substantially, so the perturbation no longer probes the labelled feature.
The conservative 0.1 threshold keeps this distortion small.
The full list and pairwise overlap are in Appendix~\ref{app:directions}.
For MELBO, we use 528 pairs obtained via the procedure described in Section~\ref{sec:methods}.
For SAE latents, we use top-activating decoder columns from Gemma Scope's residual stream layer 2 width-16k SAE.
PCA and random directions serve as non-feature baselines, as do random-difference directions, for which we form 780 pairs from 40 directions and keep the 710 with intra-pair cosine similarity below 0.1.
\begin{figure*}[!t]
  \centering
  \includegraphics[width=0.92\textwidth]{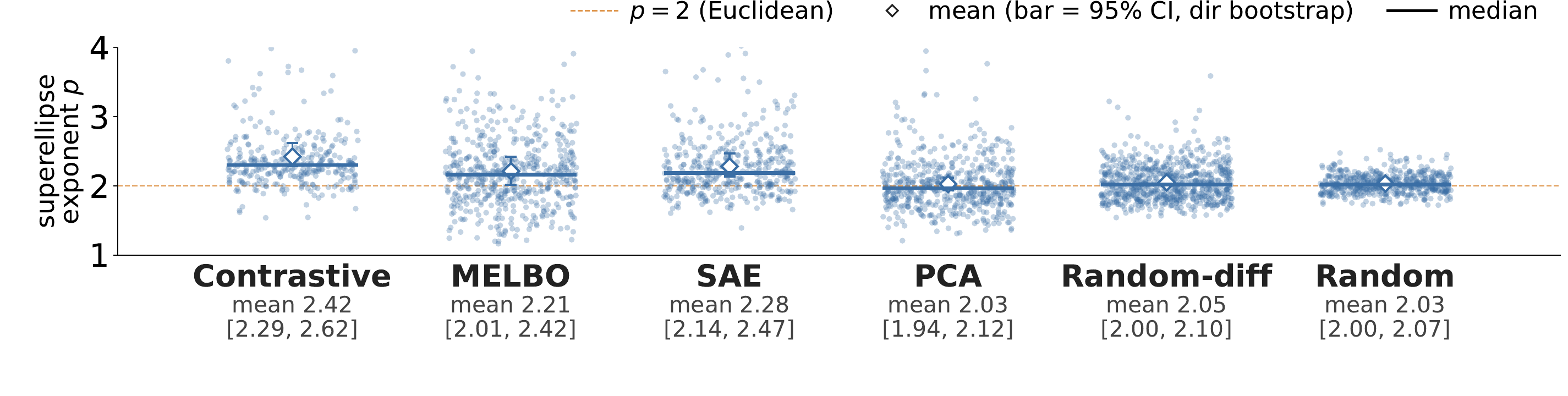}
  \caption{\textbf{Superellipse exponents by direction type.}
    Each dot is one fitted superellipse exponent $p$ for a pair of directions.
    The black horizontal lines are the per-column medians.
    The dashed orange line corresponds to the $p=2$ isotropic reference.
    The white markers are the per-column means, with error bars corresponding to the 95\% confidence interval on the mean, computed by direction bootstrapping.
    Candidate feature directions (contrastive, MELBO, top-activating SAE latents) sit
    consistently above $p=2$.
    The PCA, random, and random-difference baselines cluster at $p\approx 2$.
    The superellipse fits are good: no per-condition median fit residual exceeds $2\%$.}
  \label{fig:beeswarm_directions}
\end{figure*}
We observe that contrastive directions (mean 2.42, 95\% CI [2.29, 2.62], median 2.30), MELBO directions (mean 2.21, 95\% CI [2.01, 2.42], median 2.16), and top-activating SAE latents (mean 2.28, 95\% CI [2.14, 2.47], median 2.19) sit consistently above $p = 2$.
This indicates that the model is more sensitive to perturbations along these directions than along their mixtures, consistent with them being features and with FSEC.
We emphasize that our result is not merely that these directions are sensitive---as prior work on activation plateaus and MELBO has shown---but that they are more sensitive than mixtures of them, which is precisely what $p > 2$ measures.
The same is not true for any of our baselines: top PCA directions (mean 2.03, 95\% CI [1.94, 2.12], median 1.97), random-difference directions (mean 2.05, 95\% CI [2.00, 2.10], median 2.02), and random directions (mean 2.03, 95\% CI [2.00, 2.07], median 2.02) all cluster at $p \approx 2$.
As discussed in Section~\ref{sec:introduction}, $p = 2$ means that, in the per-axis-normalised coordinates, the plateau boundary is elliptical: mixtures break through at the same rescaled magnitude as the pure directions, so neither direction is privileged over its mixtures.
This is inconsistent with the baseline directions being aligned with feature directions, under the assumption of FSEC.

\subsection{Rotating away from feature directions}
\label{sec:misalignment}
\begin{figure}[!t]
  \centering
  \includegraphics[width=\columnwidth]{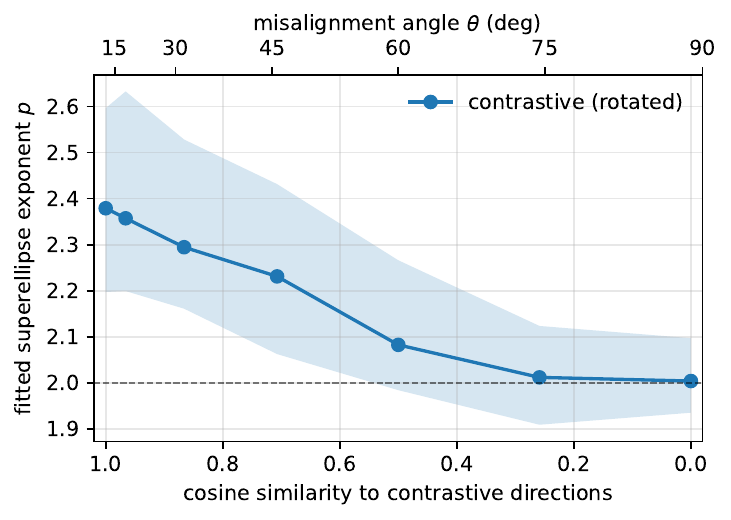}
  \caption{\textbf{Superellipse exponent versus rotation away from contrastive directions.}
    Each point is the median fitted $p$ over the 40 overlap-filtered contrastive pairs and four independent rotation realizations per pair; the shaded band shows the interquartile range.
    The dashed line marks the $p = 2$ isotropic reference.
    At $\cos\theta = 1$ the perturbation directions are the contrastive directions; the median there ($p \approx 2.4$) is slightly above the full contrastive-set median ($2.30$, Figure~\ref{fig:beeswarm_directions}) as it is computed over a 40-pair subsample.
    At $\cos\theta = 0$ they have been rotated fully onto a random orthogonal direction.
    $p$ degrades monotonically from $\approx 2.4$ toward $2$ as the directions are rotated away from the contrastive features.}
  \label{fig:misalignment_llm}
\end{figure}

We have shown that candidate feature directions have $p>2$ while controls do not.
We now show that $p$ decays monotonically to $p=2$ as we rotate away from candidate feature directions.
We take a pair of contrastive feature directions and we rotate each of the directions toward a fixed random orthogonal direction by an angle $\theta$, re-orthonormalize the pair, and refit, so that $\theta = 0$ recovers the original contrastive pair and $\theta = \pi/2$ yields a random direction.
We sweep $\theta \in [0, \pi/2]$ over a subsample of 40 overlap-filtered pairs (subsampled only to keep the computational cost manageable, as the full sweep is more expensive than the single-pair fits of Section~\ref{sec:exponents}). For each pair we average over four independent rotation realizations: within each, every direction is tilted toward its own fixed random orthogonal target (Figure~\ref{fig:misalignment_llm}).
The fitted exponent decreases monotonically from $p \approx 2.4$ at the contrastive directions to $p \approx 2.0$ once they are fully rotated to random, indicating that $p$ tracks alignment with the candidate feature directions.
Because we lack ground-truth features in the LLM, this sweep can only rotate away from candidate features.
In Section~\ref{sec:toy_model} we close this gap in a toy model with known features.

\subsection{Ablations}
\label{sec:ablations}
We now show that our results hold across a variety of settings.
To do so, we vary our setup along the following axes:
\begin{itemize}
  \item \textbf{Perturbation method:} additive rather than norm-matched.
  \item \textbf{Perturbation layer:} $\ell \in \{5, 10, 20\}$ in addition to the default $\ell=2$.
  \item \textbf{Measurement layer:} layers 30 and 35 in addition to the default second-to-last (layer 40 for Gemma-2-9B).
  \item \textbf{Response metric:} cosine distance at second-to-last residual stream layer and KL-divergence in logit space in addition to $L^2$ distance.
  \item \textbf{Plateau-breaking threshold:} 50\% and 200\% of the per-pair threshold $T$.
  \item \textbf{Activation source:} wikipedia-en, wikipedia-zh, and the-stack (default FineWeb).
  \item \textbf{Model:} Qwen3-1.7B, Llama-3.1-8B, Mistral-7B-v0.3, Aya-Expanse-8B, and Yi-1.5-9B.
\end{itemize}
\begin{figure*}[!t]
  \centering
  \includegraphics[width=0.92\textwidth]{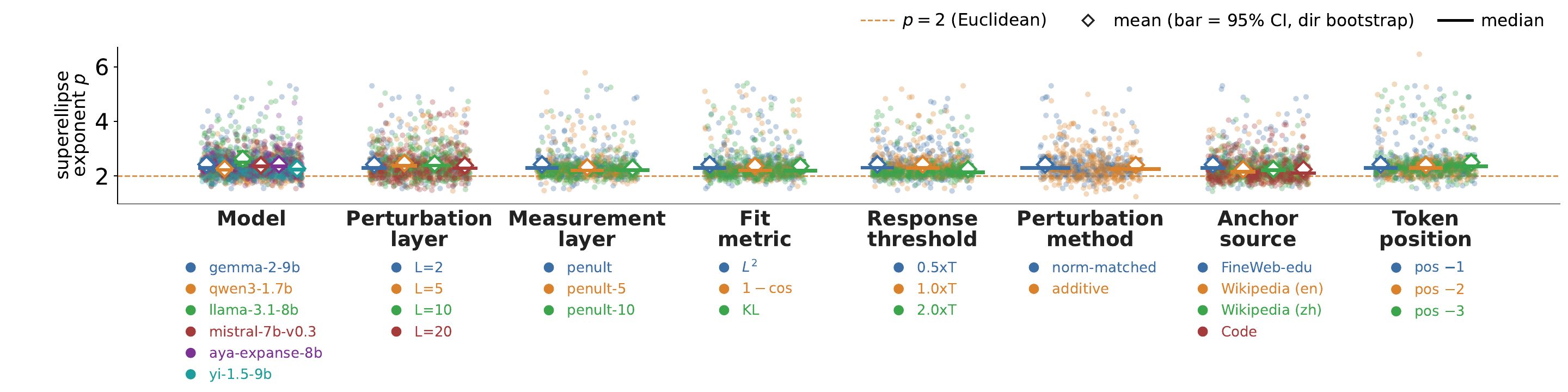}
  \caption{\textbf{Robustness of $p > 2$ for contrastive directions across setup choices.}
    Each dot is one fitted superellipse exponent $p$ for a pair of contrastive directions;
    the colored marker shows the per-setting mean with 95\% CI, and the horizontal dash shows the median.
    The dashed orange line is the $p=2$ isotropic reference.
    Columns vary the model, perturbation layer, measurement layer, response metric, response threshold, perturbation method, activation source, and perturbation token position.
    Fitted exponents remain above $p = 2$ across all variations.
    The superellipse fits are good: no per-condition median fit residual exceeds $2\%$.}
  \label{fig:beeswarm}
\end{figure*}
All our results hold across every ablation; the full results are shown in Figure~\ref{fig:beeswarm}.
For contrastive feature directions, the means are all in the [2.24, 2.63] interval, and the medians in [2.11, 2.50], all consistent with contrastive feature directions being privileged.

\section{Toy model validation}
\label{sec:toy_model}

To validate our method, we test whether it detects FSEC in a toy model with known ground-truth features.
This lets us (i) verify that our methodology yields high $p$ and good superellipse fits on known ground-truth features, and (ii) repeat the misalignment sweep of Section~\ref{sec:misalignment} from a known starting point---rotating away from the \emph{true} features rather than from imperfect candidates, since the toy model exposes the true feature directions.
Because our candidate feature directions in LLMs are imperfect approximations of the true features, we expect the measured $p$ to be biased downward, since rotating away from a true feature direction should yield a less privileged direction.

We use the two-layer denoising network of~\citet{vaintrob2026tale} as our toy model.
It denoises two-hot $d$-dimensional inputs corrupted with Gaussian noise using $H < d$ neurons, so the $d$ features are denoised (i.e., computed) in superposition over fewer than $d$ dimensions.
The input is expressed in the feature basis, giving us direct access to the ground-truth feature directions.

We apply our two-direction perturbation analysis as before, perturbing additively along pairs of ground-truth inactive feature directions.
To measure the effect of misalignment, we rotate the perturbation directions away from the true feature directions toward random directions by varying amounts.
We sweep three feature-to-neuron ratios ($4\times$, $8\times$, $16\times$) with $H = 1024$ neurons.
Full details of the toy model setup are in Appendix~\ref{app:toy_model}.

\begin{figure}[!t]
  \centering
  \includegraphics[width=\columnwidth]{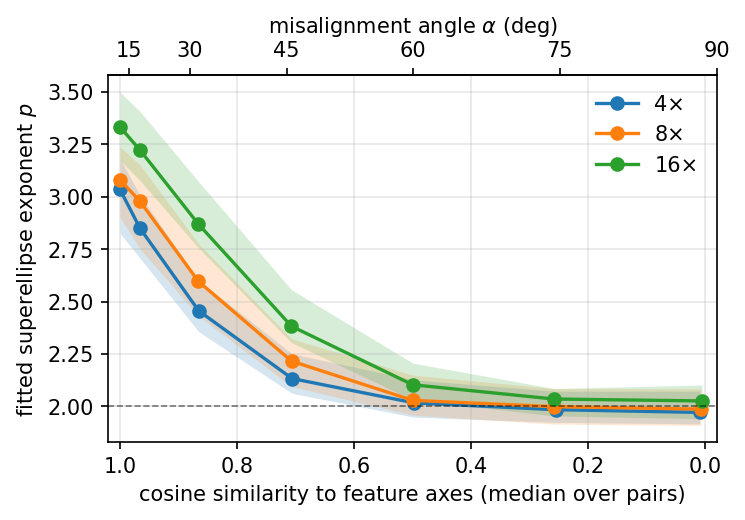}
  \caption{\textbf{Superellipse exponents versus alignment with feature directions.}
    Each dot is the median of $80$ fitted superellipse exponents; the shaded region shows the interquartile range.
    The dashed line marks the $p=2$ isotropic reference.
    Perturbation directions take the form $\mathbf{u} \propto \cos\theta\,\mathbf{e}_j + \sin\theta\,\mathbf{w}$ (normalized to unit length) for $\theta \in [0, \pi/2]$, where $\mathbf{e}_j$ is a feature axis and $\mathbf{w}$ is an isotropic random unit vector in $\mathbb{R}^d$; the x-axis shows the median realized $|\langle \mathbf{u}, \mathbf{e}_j\rangle|$ over the 80 pairs.
    The leftmost points correspond to ground-truth feature directions; the rightmost to isotropic random unit vectors, whose realized overlap with the feature axes is of order $1/\sqrt{d}$ per pair.      The three lines correspond to feature-to-neuron ratios of $4\times$, $8\times$, and $16\times$, all with $H=1024$ neurons. Ground-truth feature directions yield $p$ in the range $3$–$3.4$; $p$ degrades monotonically toward $2$ as directions are rotated away from the true features.}
  \label{fig:p_vs_cosine}
\end{figure}

Figure~\ref{fig:p_vs_cosine} shows the fitted exponent $p$ as a function of the perturbation directions' alignment with the true feature axes.
For ground-truth feature directions, we obtain $p$ in the range $3$--$3.4$, well above $2$, increasing with the feature-to-neuron ratio (highest for $16\times$).
As directions are rotated away from the true features, $p$ degrades monotonically toward $2$, which is recovered for random directions.
This mirrors the LLM misalignment sweep (Section~\ref{sec:misalignment}), but now starting from ground-truth features rather than candidates.
This is consistent with our prediction that FSEC produces $p > 2$ and that imperfect feature alignment biases $p$ downward.
Due to the significant differences between this toy model and LLMs (two layers, binary features, $\tanh^3$ activation function), we do not place much weight on the specific values obtained; we view this primarily as validation of the methodology and the directional privileging prediction.

\section{Discussion}
\label{sec:discussion}
\subsection{Why \texorpdfstring{$p > 2$}{p > 2} is necessary for error correction}
\label{sec:why-p-gt-2}

When neural networks use superposition, features are embedded non-orthogonally, so each active feature produces a small spurious ``interference'' activation along the others.
To compute in superposition despite this interference noise, a network must implement error correction~\cite{hanni2024mathematical}, and such a mechanism must treat feature directions differently from non-feature directions.
To test whether this is the case, we measure the exponent $p$ in our response model (Equation~\ref{eq:response-model}).

The exponent $p$ controls how selectively the network can respond to individual feature directions.
When $p = 2$, the response reduces to a quadratic form,
\begin{equation}
R(\hat{\mathbf{d}}) = F\!\left(\hat{\mathbf{d}}^\top \mathbf{A}\, \hat{\mathbf{d}}\right),
\end{equation}
where $\mathbf{A} = \sum_i w_i \hat{\mathbf{f}}_i \hat{\mathbf{f}}_i^\top$ is a positive semidefinite matrix with at most $d_{\text{model}}$ nonzero eigenvalues.
Under superposition, we expect far more features than dimensions, yet a quadratic form has only $d_{\text{model}}$ degrees of freedom and therefore cannot privilege that many directions above all others.
Even without superposition, $p = 2$ does not allow all feature directions to be more responsive than all non-feature directions: if the eigenvalues of $\mathbf{A}$ are not all equal, convexity implies that some non-feature directions (linear combinations of high-eigenvalue features) will be more responsive than the lowest-eigenvalue features; if all eigenvalues are equal, the response is isotropic and no direction is privileged at all.
In either case, $p = 2$ is insufficient for error correction that privileges feature directions.

At the other extreme, $p \to \infty$ makes the response sensitive only to the single largest feature projection, perfectly suppressing interference from all other directions.
This extreme is likely not how LLMs function either.
We therefore expect the true exponent to lie somewhere between these extremes: $2 < p < \infty$.

Empirically, we find $p \approx 2.3$ for contrastive directions and $p \approx 2.2$ for MELBO and SAE directions, consistently above $p = 2$ and consistent with FSEC.
Since our candidate feature directions are imperfect approximations of the model's true features, the measured $p$ is likely a lower bound on the exponent that would be obtained with ground-truth feature directions: imperfect alignment would dilute the directional sensitivity, biasing $p$ toward $2$.
Our toy model analysis (Section~\ref{sec:toy_model}) supports this, showing that $p$ degrades monotonically toward $2$ as directions are rotated away from the true features.
In fact, moderate misalignment of perturbation directions in the form of $\sim0.8$ cosine similarity with the feature directions reduces the fitted superellipse exponent $p$ from $3$--$3.4$ to $2.3$--$2.6$, indicating that our measured $p\approx2.3$ for contrastive directions in LLMs is consistent with significantly higher $p$ for true feature directions. 

\subsection{Alternative interpretations}
\label{sec:alternatives}

Our finding that $p > 2$ for candidate feature directions is consistent with FSEC.
However, there exist alternative interpretations of this result and of activation plateaus more broadly.

Activation plateaus, the near-zero downstream response to small perturbations that underlies our measurement, might not be signatures of error correction at all.
They could instead reflect learned robustness, perhaps related to broad basins of attraction in the activation space of LLMs.
This could hold whether or not LLMs implement error correction: it could be that error correction is present but plateaus are not the right empirical signature, or that error correction is absent and plateaus reflect a different phenomenon.
However, our main result does not depend on interpreting plateaus as error correction per se: even if the mechanism underlying plateaus is not error correction, the finding that $p > 2$ for candidate feature directions but $p \approx 2$ for baselines remains, and still indicates that the network treats these directions preferentially.

It is also possible that LLMs implement error correction in a way that is not directly direction-aware, i.e., without features being linear directions that are consistent across the activation space.
For instance, the in-distribution activations of LLMs may lie on a lower-dimensional manifold~\cite{bengio2017deep,cheng2023bridging}, and activation plateaus could reflect a mechanism by which small off-manifold perturbations are mapped back on-manifold.
This would be broadly consistent with our observations but would admit a different interpretation than the one we have put forward.

\section{Conclusion}
\label{sec:conclusion}
We have proposed perturbation analysis of residual-stream activations as an empirical test of whether feature directions are privileged and whether LLMs implement error correction.
We find that candidate feature directions in the form of contrastive directions, MELBO directions, and top-activating SAE latents are privileged, consistent with superposition and error correction.
Our results hold across six model families and a wide range of experimental settings.

Our results consistently select $p \approx 2.3$ for our best candidates for feature directions.
As discussed in Section~\ref{sec:discussion}, this is likely a lower bound: imperfect alignment between our candidate feature directions and the model's true features would dilute the directional sensitivity,
biasing $p$ toward $2$. This provides a quantitative target for toy models of error correction: a good toy model should reproduce $p > 2$ under comparable experimental conditions.
Indeed, the toy model of Section~\ref{sec:toy_model} passes this test, yielding $p$ in the range $3$--$3.4$ for ground-truth features.
More broadly, the exponent cleanly separates candidate feature directions ($p > 2$) from baselines ($p \approx 2$).

\paragraph{Future work.}
This discriminative power suggests using the exponent directly as an unsupervised feature-finding objective: searching for directions that maximize it.
We have explored this in preliminary experiments, and our tentative conclusion is that it does not by itself recover features: $p > 2$ appears to be necessary but not sufficient for a direction to be a feature, so maximizing the exponent alone surfaces high-$p$ directions that do not otherwise behave like features.
We flag this to avoid overstating the objective's promise; these results are preliminary and not reported here, and we leave a fuller treatment to future work.
The toy model offers a controlled test of this idea, since its ground-truth features let us check whether maximizing the exponent recovers them.
On the modeling side, extending the toy-model validation to standard activation functions and deeper architectures would test how closely a toy model of error correction can reproduce the exponents we measure in LLMs.

\section{Acknowledgements}
We thank Dmitry Vaintrob for pointing us to the denoising toy model and for useful discussions.
\bibliography{paper}

\begin{thebibliography}{28}
\providecommand{\natexlab}[1]{#1}
\providecommand{\url}[1]{\texttt{#1}}
\expandafter\ifx\csname urlstyle\endcsname\relax
  \providecommand{\doi}[1]{doi: #1}\else
  \providecommand{\doi}{doi: \begingroup \urlstyle{rm}\Url}\fi

\bibitem[Adler \& Shavit(2024)Adler and Shavit]{adler2024complexity}
Adler, M. and Shavit, N.
\newblock On the complexity of neural computation in superposition.
\newblock \emph{arXiv preprint arXiv:2409.15318}, 2024.

\bibitem[Bricken et~al.(2023)Bricken, Templeton, Batson, Chen, Jermyn, Conerly,
  Turner, Anil, Denison, Askell, Lasenby, Wu, Kravec, Schiefer, Maxwell,
  Joseph, Hatfield-Dodds, Tamkin, Nguyen, McLean, Burke, Hume, Carter,
  Henighan, and Olah]{bricken2023monosemanticity}
Bricken, T., Templeton, A., Batson, J., Chen, B., Jermyn, A., Conerly, T.,
  Turner, N.~L., Anil, C., Denison, C., Askell, A., Lasenby, R., Wu, Y.,
  Kravec, S., Schiefer, N., Maxwell, T., Joseph, N., Hatfield-Dodds, Z.,
  Tamkin, A., Nguyen, K., McLean, B., Burke, J.~E., Hume, T., Carter, S.,
  Henighan, T., and Olah, C.
\newblock Towards monosemanticity: Decomposing language models with dictionary
  learning.
\newblock \emph{Transformer Circuits Thread}, 2023.
\newblock URL
  \url{https://transformer-circuits.pub/2023/monosemantic-features/index.html}.

\bibitem[Cheng et~al.(2023)Cheng, Kervadec, and Baroni]{cheng2023bridging}
Cheng, E., Kervadec, C., and Baroni, M.
\newblock Bridging information-theoretic and geometric compression in language
  models.
\newblock In \emph{Proceedings of the 2023 Conference on Empirical Methods in
  Natural Language Processing}, pp.\  12397--12420, 2023.

\bibitem[Cunningham et~al.(2023)Cunningham, Ewart, Riggs, Huben, and
  Sharkey]{cunningham2023sparse}
Cunningham, H., Ewart, A., Riggs, L., Huben, R., and Sharkey, L.
\newblock Sparse autoencoders find highly interpretable features in language
  models.
\newblock \emph{arXiv preprint arXiv:2309.08600}, 2023.

\bibitem[Dang et~al.(2024)Dang, Singh, D'souza, Ahmadian, Salamanca, Smith,
  Peppin, Hong, Govindassamy, Zhao, et~al.]{dang2024aya}
Dang, J., Singh, S., D'souza, D., Ahmadian, A., Salamanca, A., Smith, M.,
  Peppin, A., Hong, S., Govindassamy, M., Zhao, T., et~al.
\newblock Aya expanse: Combining research breakthroughs for a new multilingual
  frontier.
\newblock \emph{arXiv preprint arXiv:2412.04261}, 2024.

\bibitem[Elhage et~al.(2022)Elhage, Hume, Olsson, Schiefer, Henighan, Kravec,
  Hatfield-Dodds, Lasenby, Drain, Chen, et~al.]{elhage2022toy}
Elhage, N., Hume, T., Olsson, C., Schiefer, N., Henighan, T., Kravec, S.,
  Hatfield-Dodds, Z., Lasenby, R., Drain, D., Chen, C., et~al.
\newblock Toy models of superposition.
\newblock \emph{arXiv preprint arXiv:2209.10652}, 2022.

\bibitem[Gao et~al.(2024)Gao, la~Tour, Tillman, Goh, Troll, Radford, Sutskever,
  Leike, and Wu]{gao2024scaling}
Gao, L., la~Tour, T.~D., Tillman, H., Goh, G., Troll, R., Radford, A.,
  Sutskever, I., Leike, J., and Wu, J.
\newblock Scaling and evaluating sparse autoencoders.
\newblock \emph{arXiv preprint arXiv:2406.04093}, 2024.

\bibitem[Goodfellow et~al.(2016)Goodfellow, Bengio, and
  Courville]{bengio2017deep}
Goodfellow, I., Bengio, Y., and Courville, A.
\newblock \emph{Deep learning}.
\newblock MIT press Cambridge, MA, USA, 2016.

\bibitem[Grattafiori et~al.(2024)Grattafiori, Dubey, Jauhri, Pandey, Kadian,
  Al-Dahle, Letman, Mathur, Schelten, Vaughan, et~al.]{grattafiori2024llama}
Grattafiori, A., Dubey, A., Jauhri, A., Pandey, A., Kadian, A., Al-Dahle, A.,
  Letman, A., Mathur, A., Schelten, A., Vaughan, A., et~al.
\newblock The llama 3 herd of models.
\newblock \emph{arXiv preprint arXiv:2407.21783}, 2024.

\bibitem[H{\"a}nni et~al.(2024)H{\"a}nni, Mendel, Vaintrob, and
  Chan]{hanni2024mathematical}
H{\"a}nni, K., Mendel, J., Vaintrob, D., and Chan, L.
\newblock Mathematical models of computation in superposition.
\newblock \emph{arXiv preprint arXiv:2408.05451}, 2024.

\bibitem[Heimersheim \& Mendel(2024)Heimersheim and
  Mendel]{heimersheim2024plateaus}
Heimersheim, S. and Mendel, J.
\newblock [interim research report] {A}ctivation plateaus \& sensitive
  directions in {GPT2}.
\newblock AI Alignment Forum, July 2024.
\newblock URL
  \url{https://www.alignmentforum.org/posts/LajDyGyiyX8DNNsuF/interim-research-report-activation-plateaus-and-sensitive-1}.
\newblock Work produced at Apollo Research.

\bibitem[Heimersheim \& Nanda(2024)Heimersheim and Nanda]{heimersheim2024use}
Heimersheim, S. and Nanda, N.
\newblock How to use and interpret activation patching.
\newblock \emph{arXiv preprint arXiv:2404.15255}, 2024.

\bibitem[Janiak et~al.(2024)Janiak, Karwowski, Mangat, Giglemiani, Petrova, and
  Heimersheim]{janiak2024characterizing}
Janiak, J., Karwowski, J., Mangat, C.~S., Giglemiani, G., Petrova, N., and
  Heimersheim, S.
\newblock Characterizing stable regions in the residual stream of llms.
\newblock \emph{arXiv preprint arXiv:2409.17113}, 2024.

\bibitem[Jiang et~al.(2023)Jiang, Sablayrolles, Mensch, Bamford, Chaplot,
  de~Las~Casas, Bressand, Lengyel, Lample, Saulnier, Lavaud, Lachaux, Stock,
  Scao, Lavril, Wang, Lacroix, and Sayed]{Jiang2023Mistral7}
Jiang, A.~Q., Sablayrolles, A., Mensch, A., Bamford, C., Chaplot, D.~S.,
  de~Las~Casas, D., Bressand, F., Lengyel, G., Lample, G., Saulnier, L.,
  Lavaud, L.~R., Lachaux, M.-A., Stock, P., Scao, T.~L., Lavril, T., Wang, T.,
  Lacroix, T., and Sayed, W.~E.
\newblock Mistral 7b.
\newblock \emph{ArXiv}, abs/2310.06825, 2023.

\bibitem[Lieberum et~al.(2024)Lieberum, Rajamanoharan, Conmy, Smith, Sonnerat,
  Varma, Kram{\'a}r, Dragan, Shah, and Nanda]{lieberum2024gemma}
Lieberum, T., Rajamanoharan, S., Conmy, A., Smith, L., Sonnerat, N., Varma, V.,
  Kram{\'a}r, J., Dragan, A., Shah, R., and Nanda, N.
\newblock Gemma scope: Open sparse autoencoders everywhere all at once on gemma
  2.
\newblock In \emph{Proceedings of the 7th BlackboxNLP Workshop: Analyzing and
  Interpreting Neural Networks for NLP}, pp.\  278--300, 2024.

\bibitem[Mack \& Turner(2024{\natexlab{a}})Mack and
  Turner]{mack2024deepcausaltranscoding}
Mack, A. and Turner, A.
\newblock Deep causal transcoding: A framework for mechanistically eliciting
  latent behaviors in language models, 12 2024{\natexlab{a}}.
\newblock URL
  \url{https://www.lesswrong.com/posts/fSRg5qs9TPbNy3sm5/deep-causal-transcoding-a-framework-for-mechanistically}.
\newblock LessWrong.

\bibitem[Mack \& Turner(2024{\natexlab{b}})Mack and Turner]{mack2024melbo}
Mack, A. and Turner, A.
\newblock Mechanistically eliciting latent behaviors in language models, 4
  2024{\natexlab{b}}.
\newblock URL
  \url{https://www.lesswrong.com/posts/ioPnHKFyy4Cw2Gr2x/mechanistically-eliciting-latent-behaviors-in-language-1}.
\newblock LessWrong.

\bibitem[Meng et~al.(2022)Meng, Bau, Andonian, and Belinkov]{meng2022locating}
Meng, K., Bau, D., Andonian, A., and Belinkov, Y.
\newblock Locating and editing factual associations in gpt.
\newblock \emph{Advances in neural information processing systems},
  35:\penalty0 17359--17372, 2022.

\bibitem[Olah et~al.(2025)Olah, Turner, and Conerly]{olah2025interference}
Olah, C., Turner, N.~L., and Conerly, T.
\newblock A toy model of interference weights.
\newblock \emph{Transformer Circuits Thread}, 2025.
\newblock URL
  \url{https://transformer-circuits.pub/2025/interference-weights/index.html}.

\bibitem[Panickssery et~al.(2023)Panickssery, Gabrieli, Schulz, Tong, Hubinger,
  and Turner]{panickssery2023steering}
Panickssery, N., Gabrieli, N., Schulz, J., Tong, M., Hubinger, E., and Turner,
  A.~M.
\newblock Steering llama 2 via contrastive activation addition.
\newblock \emph{arXiv preprint arXiv:2312.06681}, 2023.

\bibitem[Penedo et~al.(2024)Penedo, Kydl{\'\i}{\v{c}}ek, Ben~allal, Lozhkov,
  Mitchell, Raffel, Von~Werra, Wolf, et~al.]{penedo2024fineweb}
Penedo, G., Kydl{\'\i}{\v{c}}ek, H., Ben~allal, L., Lozhkov, A., Mitchell, M.,
  Raffel, C., Von~Werra, L., Wolf, T., et~al.
\newblock The fineweb datasets: Decanting the web for the finest text data at
  scale.
\newblock \emph{Advances in Neural Information Processing Systems},
  37:\penalty0 30811--30849, 2024.

\bibitem[Shinkle \& Heimersheim(2025)Shinkle and
  Heimersheim]{shinkle2025plateaus}
Shinkle, M. and Heimersheim, S.
\newblock Activation plateaus: Where and how they emerge.
\newblock LessWrong, October 2025.
\newblock URL
  \url{https://www.lesswrong.com/posts/WMfSbt7AAcJdHzysB/activation-plateaus-where-and-how-they-emerge}.
\newblock Accessed: 2026-04-27.

\bibitem[Team et~al.(2024)Team, Riviere, Pathak, Sessa, Hardin, Bhupatiraju,
  Hussenot, Mesnard, Shahriari, Ram{\'e}, et~al.]{team2024gemma}
Team, G., Riviere, M., Pathak, S., Sessa, P.~G., Hardin, C., Bhupatiraju, S.,
  Hussenot, L., Mesnard, T., Shahriari, B., Ram{\'e}, A., et~al.
\newblock Gemma 2: Improving open language models at a practical size.
\newblock \emph{arXiv preprint arXiv:2408.00118}, 2024.

\bibitem[Templeton et~al.(2026)Templeton, Conerly, Marcus, Lindsey, Bricken,
  Chen, Pearce, Citro, Ameisen, Jones, et~al.]{templeton2026scaling}
Templeton, A., Conerly, T., Marcus, J., Lindsey, J., Bricken, T., Chen, B.,
  Pearce, A., Citro, C., Ameisen, E., Jones, A., et~al.
\newblock Scaling monosemanticity: Extracting interpretable features from
  claude 3 sonnet.
\newblock \emph{arXiv preprint arXiv:2605.29358}, 2026.

\bibitem[Turner et~al.(2023)Turner, Thiergart, Leech, Udell, Vazquez, Mini, and
  MacDiarmid]{turner2023steering}
Turner, A.~M., Thiergart, L., Leech, G., Udell, D., Vazquez, J.~J., Mini, U.,
  and MacDiarmid, M.
\newblock Steering language models with activation engineering.
\newblock \emph{arXiv preprint arXiv:2308.10248}, 2023.

\bibitem[Vaintrob(2026)]{vaintrob2026tale}
Vaintrob, D.
\newblock A tale of three theories: Sparsity, frustration, and statistical
  field theory.
\newblock LessWrong, January 2026.
\newblock URL
  \url{https://www.lesswrong.com/posts/siu22scEfuKxpSgfK/a-tale-of-three-theories-sparsity-frustration-and}.
\newblock Accessed: 2026-05-06.

\bibitem[Yang et~al.(2025)Yang, Li, Yang, Zhang, Hui, Zheng, Yu, Gao, Huang,
  Lv, et~al.]{yang2025qwen3}
Yang, A., Li, A., Yang, B., Zhang, B., Hui, B., Zheng, B., Yu, B., Gao, C.,
  Huang, C., Lv, C., et~al.
\newblock Qwen3 technical report.
\newblock \emph{arXiv preprint arXiv:2505.09388}, 2025.

\bibitem[Young et~al.(2024)Young, Chen, Li, Huang, Zhang, Zhang, Wang, Li, Zhu,
  Chen, et~al.]{young2024yi}
Young, A., Chen, B., Li, C., Huang, C., Zhang, G., Zhang, G., Wang, G., Li, H.,
  Zhu, J., Chen, J., et~al.
\newblock Yi: Open foundation models by 01. ai.
\newblock \emph{arXiv preprint arXiv:2403.04652}, 2024.

\end{thebibliography}
\bibliographystyle{icml2026}

\appendix

\section{Contrastive direction inventory}
\label{app:directions}
We extract 33 contrastive directions across three families.
Each direction is the difference of mean last-token activations at the perturbation layer between a positive prompt set and a matched negative set, normalised to a unit vector.

The 14 \textbf{binary semantic} directions are Age, Certainty, Era, Gender, Health, Honesty, Literary, Number, Person, Refusal, Sentiment, Status, Tense, and Wealth.
Each uses 30 matched prompt pairs except Literary, which contrasts the slang and literary registers across 50 prompt templates.
Prompts are LLM-generated and human-verified.

The 10 \textbf{natural-language} directions cover Arabic, Chinese, Dutch, French, German, Italian, Japanese, Portuguese, Russian, and Spanish.
Each pair is an English prompt and its translation in the target language; we use 36 matched prompts per direction, drawn from a single multilingual prompt pool so that all 10 directions share the same English source side.

The 9 \textbf{programming-language} directions cover C++, Go, Haskell, Java, JavaScript, Lisp, Python, Rust, and TypeScript.
Each pair is an English prompt and an equivalent code snippet in the target language; 30 matched prompts per direction.

\begin{figure*}[!ht]
  \centering
  \includegraphics[width=0.78\textwidth]{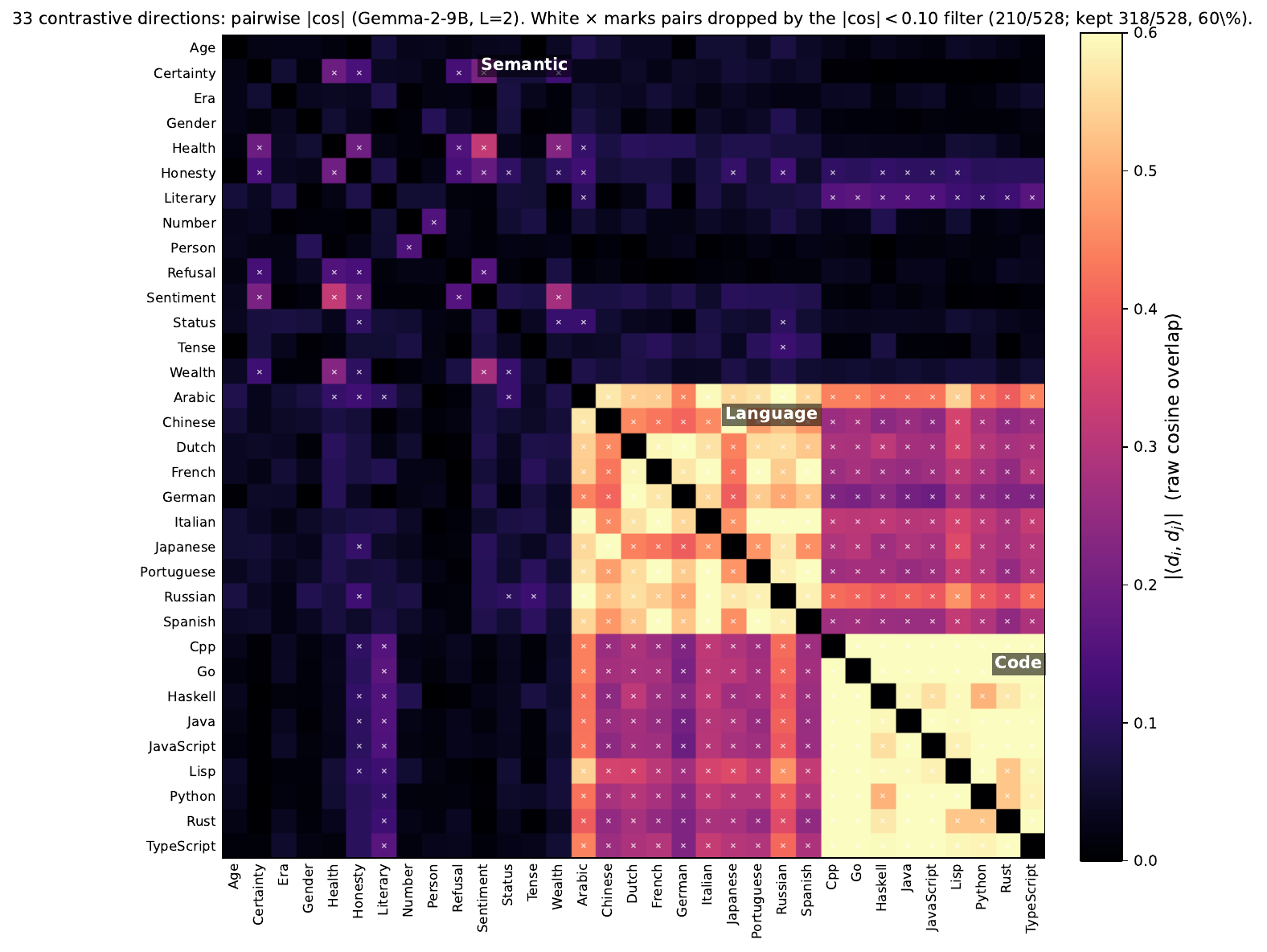}
  \caption{\textbf{Pairwise cosine overlap of the 33 contrastive directions.}
    Cells show $|\langle \mathbf{d}_i, \mathbf{d}_j\rangle|$ on Gemma-2-9B at layer~2.
    Directions are ordered and labelled by family (semantic, natural language, programming language).
    White $\times$ marks the 210 pairs dropped by the $|\cos|<0.1$ filter applied throughout the paper; the remaining 318 pairs are the ones plotted as Contrastive in the beeswarms.
    Most filtered pairs sit inside the natural-language block.}
  \label{fig:directions_overlap}
\end{figure*}

Forming all $\binom{33}{2}=528$ unordered pairs, we keep the 318 with raw overlap $|\langle \mathbf{d}_i, \mathbf{d}_j\rangle|<0.1$ for our two-direction analysis (Figure~\ref{fig:directions_overlap}).
The filter is needed because our two-direction analysis orthogonalises each pair via Gram--Schmidt (Section~\ref{sec:methods}): if two directions overlap significantly, removing the shared component leaves only a small residual, which renormalisation to unit length then amplifies, so the resulting direction reflects this residual rather than the labelled feature.
The $0.1$ threshold is conservative---at this overlap the orthogonalised direction remains more than $99\%$ aligned with the original, and the distortion grows with overlap.

\section{Distribution of superellipse fit residuals}
\label{app:residuals}
For every fitted pair we record the mean radial fraction
\(\rho = \tfrac{1}{M}\sum_{k=1}^{M} \bigl|\,(|x_k|^p+|y_k|^p)^{1/p}-1\,\bigr|\),
i.e.\ the average fractional distance of the $M$ contour points from the fitted superellipse, with $\rho=0$ corresponding to a perfect fit.
The number of contour points $M$ varies per fit (it is the cardinality of the iso-threshold level set on the 2D $L^2$ grid); the per-fit median is $33$ (IQR $29$--$37$).

\begin{figure*}[!ht]
  \centering
  \includegraphics[width=0.95\textwidth]{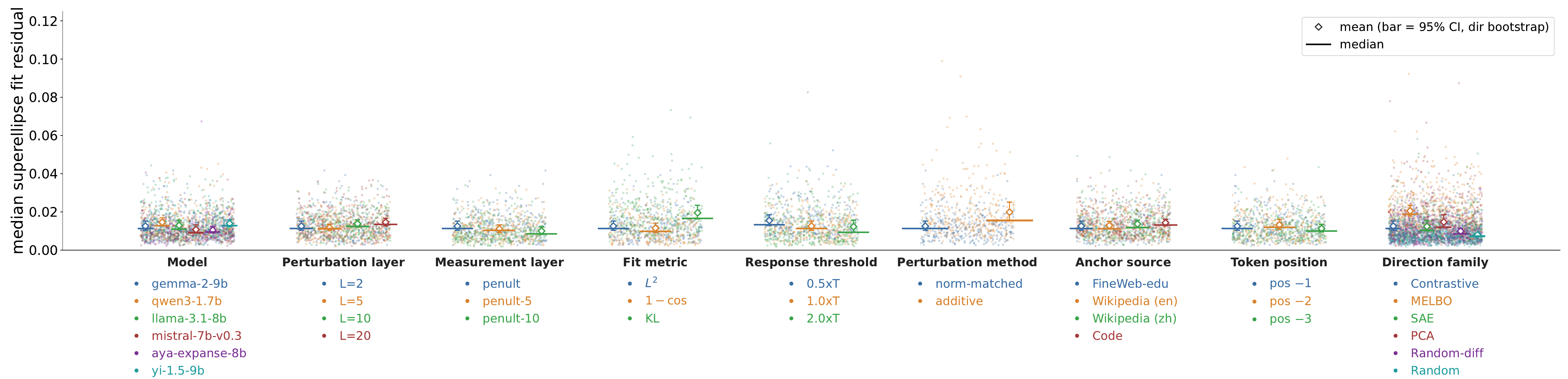}
  \caption{\textbf{Distribution of superellipse fit residuals across all conditions.}
    Each dot is one pair's fit residual $\rho$.
    The eight left-most columns mirror the ablation axes of Figure~\ref{fig:beeswarm} (Model, Perturbation layer, Measurement layer, Response metric, Response threshold, Perturbation method, Token position, Activation source).
    The right-most column gives the residual distribution per direction family (Contrastive, MELBO, SAE, PCA, Random).
    Per-sub-group mean (white diamond) $\pm$ 95\% direction-bootstrap CI and median (horizontal bar) are colour-matched to the sub-group's dots.
    The y-axis is clipped at 12.5\% to keep the bulk of the distribution legible; this clips 4 of the 10{,}719 plotted dots ($<$0.05\%).}
  \label{fig:residuals_beeswarm}
\end{figure*}

Each pair contributes one dot in Figure~\ref{fig:residuals_beeswarm}.

Every sub-group median residual is below 2\%.

\section{Toy model details}
\label{app:toy_model}

\subsection{Architecture}

The toy model follows~\citet{vaintrob2026tale}.
It is a two-layer network with tied weights: an encoder $E \in \mathbb{R}^{H \times d}$ and decoder $D = E^\top$, with a $\tanh^3$ activation function applied element-wise to the hidden layer.
The entries of $E$ are drawn i.i.d.\ from $\{0, +1, -1\}$ with probabilities $\{1-q, q/2, q/2\}$, where $q$ is a sparsity parameter controlling the fraction of non-zero entries.
The network receives a noisy two-hot input $x = x_{\text{clean}} + \epsilon$, where $x_{\text{clean}}$ has exactly $s = 2$ active coordinates set to $1$ and the remaining set to $0$, and $\epsilon \sim \mathcal{N}(0, \sigma^2 I)$ with $\sigma^2 = 0.03$.
The network operates in superposition, with the number of features $d$ exceeding the number of neurons $H$.

\subsection{Optimization}

We do not train the encoder and decoder via gradient descent.
Instead, we fix the random encoder structure and optimize two scalar parameters: an input gain $c_{\text{in}}$ and an output scale $c_{\text{out}}$, so that the network computes $y = c_{\text{out}} \cdot \tanh^3(c_{\text{in}} \cdot E x)$.
We minimize a per-coordinate weighted MSE loss:
\begin{equation}
    \mathcal{L} = \frac{1}{B \cdot d} \left[ \lambda \sum_{i \in \text{on}} (c_{\text{out}} f_i - 1)^2 + \sum_{i \in \text{off}} (c_{\text{out}} f_i)^2 \right],
\end{equation}
where $f = D \tanh^3(c_{\text{in}} \cdot E x)$, ``on'' denotes coordinates where $x_{\text{clean}} = 1$, and ``off'' denotes coordinates where $x_{\text{clean}} = 0$.
The weighting $\lambda$ upweights active-coordinate residuals to discourage the trivial solution of outputting zero, which otherwise dominates for very sparse inputs.
Given $c_{\text{in}}$, the optimal $c_{\text{out}}$ has a closed-form solution:
\begin{equation}
    c_{\text{out}}^* = \frac{\lambda \cdot \langle f, x_{\text{clean}} \rangle_{\text{on}}}{\lambda \cdot \|f\|^2_{\text{on}} + \|f\|^2_{\text{off}}}.
\end{equation}
We sweep $c_{\text{in}}$ over a 121-point grid on $[0.05, 8.0]$ and select the value minimizing $\mathcal{L}$.

The sparsity parameter $q$ was swept over $\{0.025, 0.05, 0.075, 0.1, 0.125, 0.15\}$ for each $(d, H)$ configuration, selecting $q^*$ by minimum training loss.
The selected values were $q^* = 0.15$ for $4\times$, $q^* = 0.10$ for $8\times$, and $q^* = 0.075$ for $16\times$.
The weighting parameter was set to $\lambda \in \{500, 1000, 2000\}$ for $d \in \{4096, 8192, 16384\}$ respectively, scaled roughly as $\lambda \approx d / (s \cdot \text{SNR}^2)$ with $s = 2$ and empirical $\text{SNR} \approx 2$.

\subsection{Perturbation procedure}

The active set is fixed to $S = \{0, 1\}$ throughout, so all perturbations are performed around the same clean input $x^* = e_0 + e_1$.
We sample $80$ pairs $(j, k)$ uniformly without replacement from the inactive coordinates $\{2, 3, \ldots, d-1\}$.
For each pair, we perturb along the two ground-truth feature directions $e_j$ and $e_k$.
To measure the effect of misalignment, we rotate $e_j$ and $e_k$ toward random directions by varying amounts.
Specifically, for each pair we sample an isotropic random unit pair $(w_u, w_v)$, mutually orthogonal but otherwise unconstrained, and interpolate between the feature direction and the random direction at each misalignment level.
The same $(w_u, w_v)$ pair is used across all misalignment levels for a given feature pair.

\subsection{Superellipse fitting}

\begin{figure}[!htpb]
  \centering
  \includegraphics[width=\columnwidth]{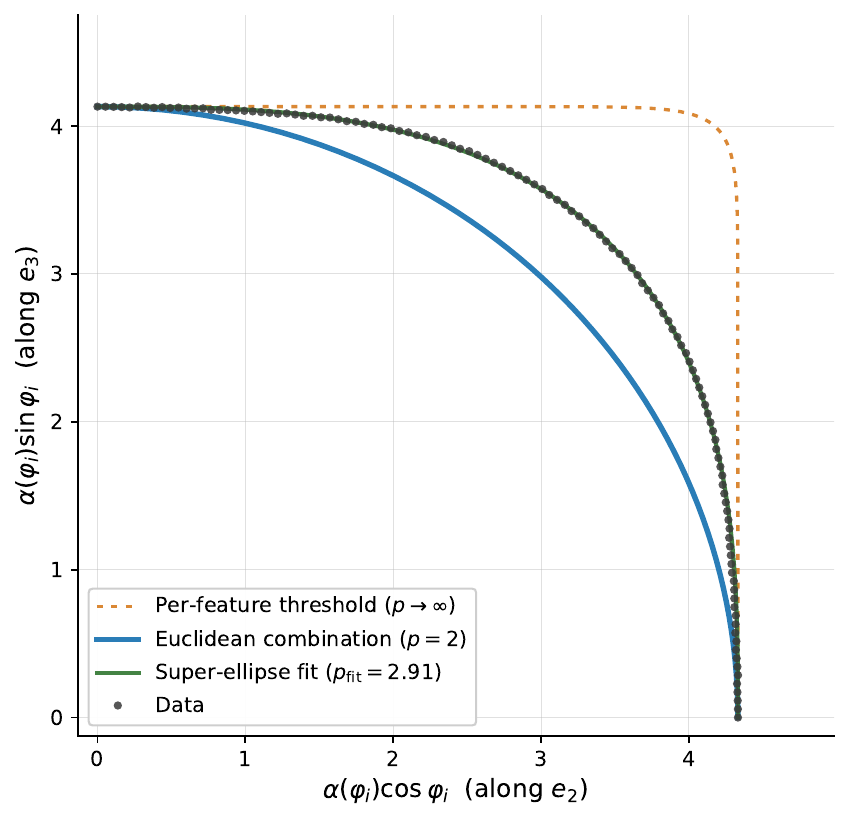}
    \caption{\textbf{Iso-plateau boundary.}
    Plateau-breaking perturbation magnitudes for perturbations aligned with combinations of two feature directions in the toy model with $d=8192$.
    The fitted superellipse exponent is $2.91$ with a mean residual of $0.13\%$.}
  \label{fig:boundary_toy}
\end{figure}

\begin{figure}[!htpb]
  \centering
  \includegraphics[width=\columnwidth]{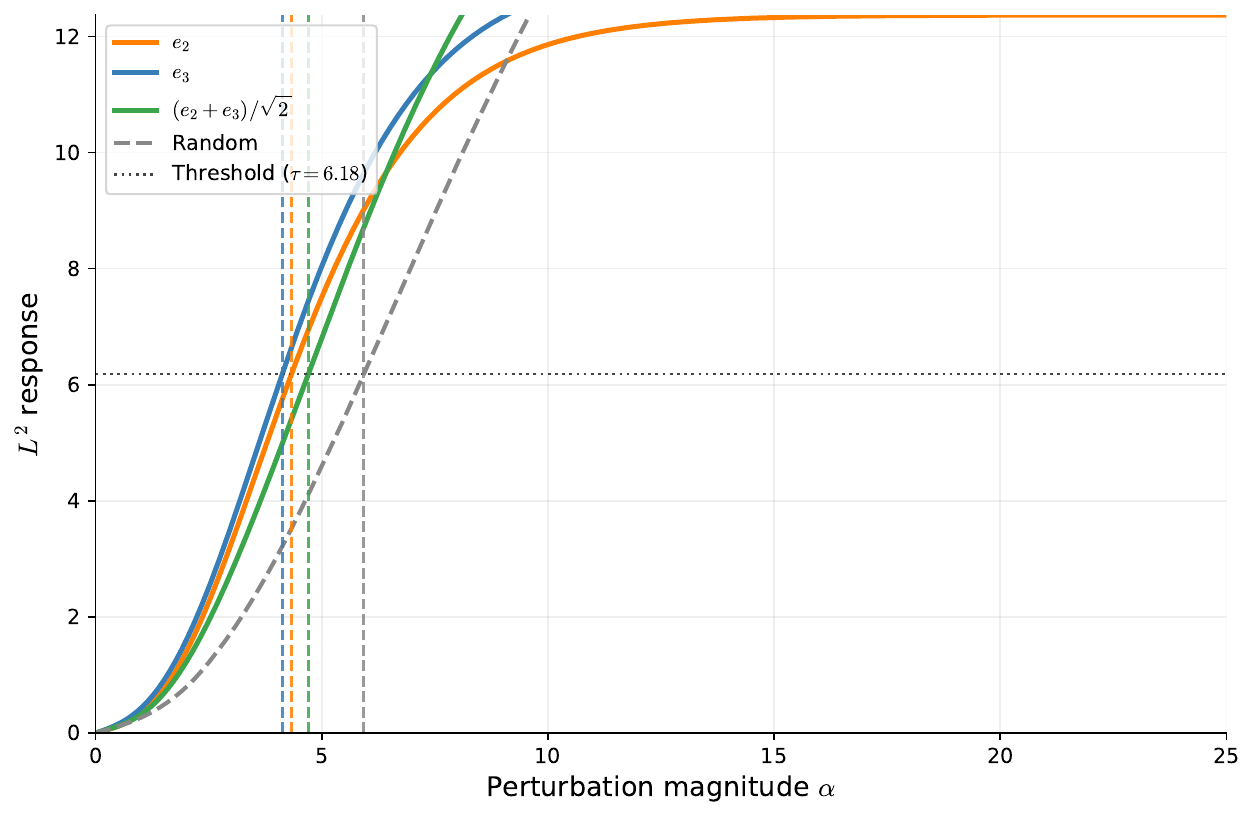}
    \caption{\textbf{Toy-model response-versus-magnitude sweep.}
    $L^2$ response of the toy model ($d=8192$, $H=1024$) to additive perturbations of magnitude $\alpha$ along two inactive feature axes $\mathbf{e}_2$ and $\mathbf{e}_3$, their equal-weight combination, and a random-direction baseline (median over 10 isotropic unit directions), with the plateau-breaking threshold $\tau$ (dotted) and each curve's plateau-breaking magnitude (dashed verticals).
    As in Figure~\ref{fig:plateau-breaking-measurement}, feature axes break the plateau first, their combination later, and random directions last.}
  \label{fig:toy_sweep}
\end{figure}

For each direction pair and misalignment level, we sweep $60$ mixing angles $\varphi$ uniformly on $[0, \pi/2]$.
For each $\varphi$, we perform a 2500-point linear scan of the perturbation magnitude $\alpha$ on $[0, 25]$ and identify the smallest $\alpha$ at which $\|f(x^* + \alpha \cdot \hat{d}(\varphi)) - f(x^*)\|_2 > \tau$.
The threshold is set to $\tau = 0.5 \cdot \max_{\alpha \leq 25} \|f(x^* + \alpha \cdot e_j) - f(x^*)\|_2$ for the first inactive feature $j = 2$, computed per network.
The superellipse exponent is estimated by minimizing $\sum_\varphi \left( (x_\varphi / r_0)^n + (y_\varphi / r_{\pi/2})^n - 1 \right)^2$ over $n \in [0.5, 60]$, where $(x_\varphi, y_\varphi) = \alpha(\varphi) \cdot (\cos\varphi, \sin\varphi)$.
The perturbation sweep is deterministic (no noise is added at perturbation time).
All random seeds are set to 42.
We show in Figure~\ref{fig:boundary_toy} an example of the superelliptical boundaries we obtain for the toy model in the style of Figure~\ref{fig:iso-plateau-boundary} in the main text.
Figure~\ref{fig:toy_sweep} shows the toy-model analogue of Figure~\ref{fig:plateau-breaking-measurement}: the $L^2$ response as a function of perturbation magnitude along two feature axes, their equal-weight combination, and a random-direction baseline.
The feature axes break the plateau first ($\alpha \approx 4.1$--$4.3$) and random directions last ($\alpha = 5.93$); the combination breaks at $\alpha = 4.70$, later than either axis, whereas $p = 2$ would predict it breaks at the same magnitude as the axes.
\end{document}